\documentclass{article}
\usepackage{ijcai20}

\usepackage{times}
\usepackage{soul}
\usepackage{url}
\usepackage[hidelinks]{hyperref}
\usepackage[utf8]{inputenc}
\usepackage[small]{caption}
\usepackage{graphicx}
\usepackage{amsmath}
\usepackage{amsthm}
\usepackage{booktabs}
\usepackage{algorithm}
\usepackage[noend]{algpseudocode}
\usepackage{researchpack}
\usepackage{wrapfig}

\newcommand{\prob}{\mathbb{P}\xspace}

\newcommand{\predy}{\ensuremath{\hat{y}}\xspace}

\newcommand{\noisyy}{\ensuremath{\tilde{y}}\xspace}

\newcommand{\consy}{\ensuremath{y'}\xspace}

\newcommand{\method}{\textsc{isgp}\xspace}
\newcommand{\acronym}{Incremental Skeptical Gaussian Processes\xspace}
\newcommand{\rf}{\textsc{srf}\xspace}
\newcommand{\gpnever}{$\textsc{gp}_\text{never}$\xspace}
\newcommand{\gpalways}{$\textsc{gp}_\text{always}$\xspace}

\title{Learning in the Wild with Incremental Skeptical Gaussian Processes}
\author{
    Andrea Bontempelli$^1$\and
    Stefano Teso$^1$\and
    Fausto Giunchiglia$^{1,2}$\And
    Andrea Passerini$^1$
    \affiliations
    $^1$University of Trento, Italy
    $^2$Jilin University, Changchun, China
    \emails
    name.surname@unitn.it
}

\begin{document}
\maketitle

\begin{abstract}

    The ability to learn from human supervision is fundamental for personal
    assistants and other interactive applications of AI.  Two central
    challenges for deploying interactive learners in the wild are the
    unreliable nature of the supervision and the varying complexity of the
    prediction task.  We address a simple but representative setting,
    \emph{incremental classification in the wild}, where the supervision is
    noisy and the number of classes grows over time.  In order to tackle this
    task, we propose a redesign of skeptical learning centered around
    Gaussian Processes (GPs).  Skeptical learning is a recent interactive
    strategy in which, if the machine is sufficiently confident that an example
    is mislabeled, it asks the annotator to reconsider her feedback.  In many
    cases, this is often enough to obtain clean supervision.  Our
    redesign, dubbed \method, leverages the uncertainty estimates supplied by GPs to better allocate labeling and contradiction queries,
    especially in the presence of noise.
    Our experiments on synthetic and real-world data show that, as a
    result, while the original formulation of skeptical learning produces over-confident models
    that can fail completely in the wild, \method works well at varying levels
    of noise and as new classes are observed.

\end{abstract}

\section{Introduction}

Imagine a handheld personal assistant that provides guidance to an end-user.
In order to give useful, timely suggestions (like ``please take your
insulin''), the agent must be aware of the user's context, for instance where
she is (``at home''), what she is doing (``eating cake''), and with whom
(``alone'').  The machine must infer this information from a stream of sensor
readings (e.g.,  GPS coordinates, nearby Bluetooth devices), with the caveat
that the target classes are user-specific (e.g., this user's home is not
another user's home) and thus that the label vocabulary must be acquired from
the user herself.  Moreover, as the user visits new places and engages in new
activities, the vocabulary changes.
This simple example shows that, in order to be successful outside of the
lab~\cite{dietterich2017steps},
AI agents must adapt to the changing conditions of the real world and to their
end-users.

We study these challenges in a simplified but non-trivial setting,
\emph{interactive classification in the wild}, where an interactive learner
requests labels from an end-user and the number of classes grows with time.
A fundamental issue in this setting is that end-users often provide unreliable
supervision~\cite{tourangeau2000psychology,west2013quality,zeni2019fixing}.
This is especially problematic in the wild, as noisy labels
may fool the machine into being under- or over-confident and into acquiring
non-existent classes.

We address these issues by proposing \acronym (\method), a redesign of
skeptical learning~\cite{zeni2019fixing} tailored for learning in the wild.  In
skeptical learning (SKL), if the interactive learner is confident that a newly
obtained example is mislabeled, it immediately asks the annotator to reconsider
her feedback.  In stark contrast to other noise handling alternatives, SKL is
designed specifically to retrieve the clean label from the annotator.


\method improves SKL in four important ways.  First,
\method builds on
Gaussian Processes (GPs)~\cite{williams2006gaussian}.  Thanks to their explicit
uncertainty estimates, GPs prevent pathological cases in which an overconfident
learner 1)~refuses to request the label of instances far from the training set,
thus failing to learn, and 2)~continuously challenges the user regardless of
her past performance, estranging her.
%
Second, \method makes use of the model's uncertainty to determine whether to be
skeptical or credulous, while SKL uses an inflexible strategy that relies on
the \emph{number} of observed examples only.
Third, while SKL relies on several hard-to-choose hyper-parameters, \method
makes use of a simple and robust algorithm that works well even without
fine-tuning.
Last, \method makes use of incremental learning techniques for improved
scalability~\cite{lutz2013want}.

Summarizing, we:
1)~Introduce interactive classification in the wild, a novel form of interactive learning in which  there is as substantial amount of labelling noise and new classes are observed over time;
2)~Develop \method, a simple and robust redesign of skeptical learning that
leverages exact uncertainty estimates to appropriately allocate queries to the
user and avoids over-confident models even in the presence of noise;
3)~Showcase the advantages of \method\ -- in terms of query budget
allocation, prediction quality and efficiency -- on a controlled
synthetic task and on a real-world task.

\begin{figure*}[tb]
    \centering
    \begin{tabular}{ccc}
        \includegraphics[height=9em]{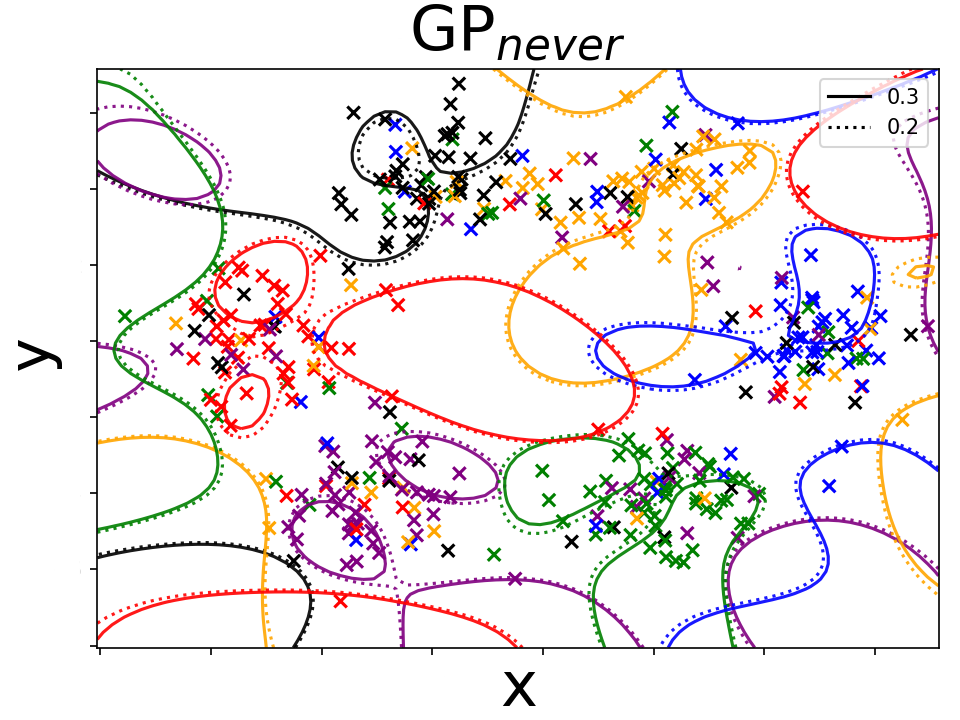} &
        \includegraphics[height=9em]{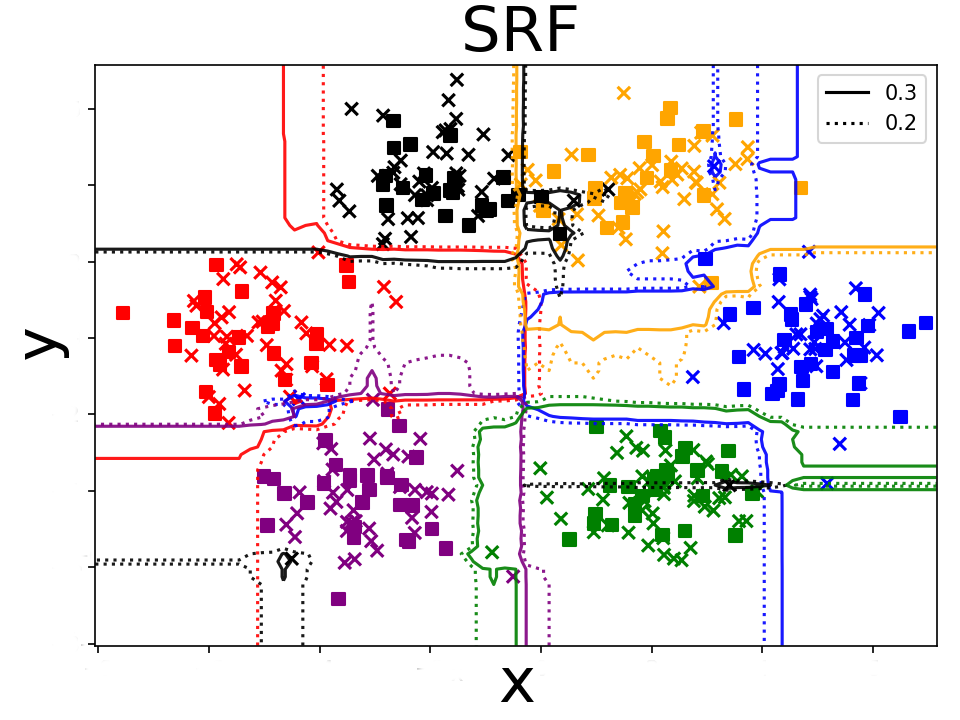} &
        \includegraphics[height=9em]{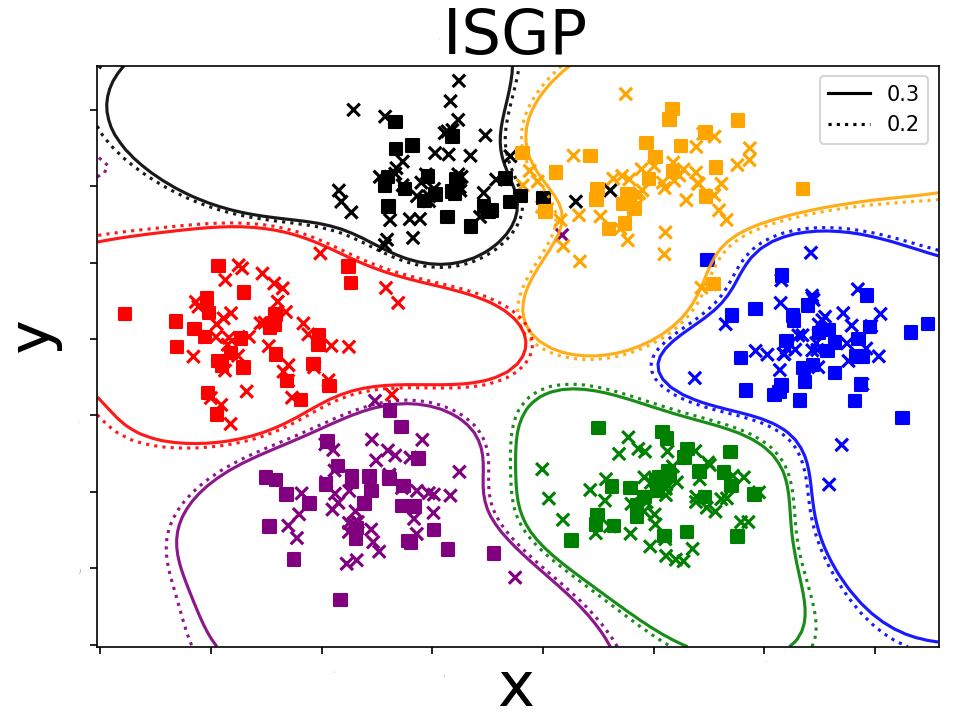}
    \end{tabular}
    \caption{\label{fig:example}  Illustration of \method on a 2D synthetic data set with six normally-distributed classes (in color) and noisy labels (corrupted at random with probability $0.4$).  The outlines enclose regions with high predictive probability (solid $\ge 0.3$, dashed $\ge
    0.2$).  Crosses and boxes are noisy examples; boxes have been cleaned by skeptical learning.   From
    the left: regular GP, skeptical learning, and \method. Best
    viewed in color.}
\end{figure*}

\section{Incremental Classification in the Wild}

Interactive classification in the wild (ICW) is a sequential prediction task:
in each round $t = 1, 2, \ldots,$ the learner receives an instance $x_t \in
\calX$ (e.g., a vector of sensor readings) and outputs a prediction $\predy_t
\in \calY$ (e.g., the user's location).  The learner is also free to query a
human annotator -- usually the end-user herself -- for the ground-truth label
$y_t \in \calY$
(e.g., the true location).  The goal of the learner is to \emph{acquire a good predictor while keeping the number of queries at a minimum}, not to overload
the annotator.

Two features make ICW unique: the amount of \emph{label noise} and the
presence of \emph{task shift}.
Label noise follows from the fact that human annotators are often subject to
momentary inattention and may fail to understand the
query~\cite{zeni2019fixing}.  The label $\noisyy_t$ fed back by the
annotator is thus often
wrong, i.e.,  $\noisyy_t \ne y_t$.  Failure to handle noise can bloat
the model and affect its accuracy~\cite{frenay2014classification}.
Label noise is especially troublesome in ICW, as it can fool the model into
being under- or over-confident.  This in turn makes it difficult to identify
informative instances and properly allocate labeling budget.

By task shift we mean that newly received instances may belong to new and
unanticipated classes.  For this reason, we distinguish between the complete
but unobserved set of classes $\calY \subseteq \bbN$ and the classes observed
up to iteration $t$, that is\footnote{It is assumed that $\calY_0$ is defined
appropriately, e.g., $|\calY_0| \ge 1$.} $\calY_t \subseteq \calY$.  Hence, $y_t$ belongs to
$\calY$, $\noisyy_t$ to $\calY_t$, and $\predy_t$ to $\calY_{t-1}$.
To keep the task manageable, we assume that previously observed classes remain
valid, i.e, $\calY_t \subseteq \calY_{t+1}$ for all $t$.  In our personal aid
example, this would imply that, for instance, the user's home remains the same
over time.  This is a reasonable assumption so long as the agent's lifetime is
not too long.  A study of other forms of task shift is left to future work.

\section{Skeptical Learning}
\label{sec:skl}

Skeptical learning (SKL) is a noise handling strategy designed for interactive
learning~\cite{zeni2019fixing}.  The idea behind SKL is simple: rather than
blindly accepting the annotator's supervision, a skeptical learner challenges
the annotator about any suspicious examples it receives.
In contrast with standard strategies for handling noise, like using robust
models or discarding anomalies~\cite{frenay2014classification}, SKL aims at
recovering the ground-truth.

In skeptical learning, an example is deemed suspicious if the learner
is confident that the model's prediction is right and that the user's
annotation is wrong.  This requires the learner to assign a confidence
level to its own predictions and to the user's annotations.  SKL
estimates these confidences using two separate heuristics.  The
confidence in the model is estimated using a combination of training
set size and confidence reported by the model.  The confidence in the
user is based on the number of user mistakes spotted during past
interaction rounds.  Both estimates are quite crude.
See~\cite{zeni2019fixing} for more details.

The original formulation of skeptical learning is not a good fit for
ICW.  First and foremost, SKL is based on random forests (RFs), which
are robust to noise but also notoriously over-confident.  This can be
clearly seen in Figure~\ref{fig:example}: the RF in the middle plot is
very confident even far away from the training set.  This is a major
issue, as over-confident predictors may stubbornly refuse to query
novel and informative instances, compromising learning, and may keep
challenging the user regardless of her past performance, overloading
and estranging her.
In addition, SKL structures learning into three stages: initially, the
machine always requests supervision and never contradicts the user;
once the model is confident enough, it begins to challenge the user;
finally, the model begins to actively request for labels.  This
partially avoids over-confidence by requesting extra labels.
This strategy fails in the wild, as new classes appear even in later
learning stages, in which over-confident models may refuse to request
supervision for them.  This occurs frequently in our experiments.
Two other issues are that SKL requires to choose several hyper-parameters (like $\theta$, which controls when to transition between stages), which is non-trivial in
interactive settings, and that it retrains the RF from scratch in each iteration.

\section{Skeptical Learning in the Wild}

\method is a redesign of skeptical learning based on Gaussian Processes (GPs) that
avoids over-confident predictors and handles label noise.  GPs are a natural choice
in learning tasks like active learning~\cite{kapoor2007active,rodrigues2014gaussian},
online bandits~\cite{srinivas2012information}, and preference elicitation~\cite{guo2010gaussian},
in which uncertainty estimates help to guide the interaction with the user.  Our
key observation is that skeptical learning is another such application.

\subsection{Gaussian Processes}

Gaussian Processes (GPs)~\cite{williams2006gaussian} are non-parametric
distributions over functions $f:\calX \to \bbR$.  A GP is entirely
specified by a mean function $\mu(x)$ and a covariance function $k(x,x')$.  The
latter encodes structural assumptions about the functions modeled by the GP
and can be implemented with any kernel function.  When no evidence is given, it
can be assumed w.l.o.g. that $\mu(x) \equiv 0$.
Bayesian inference, that is, conditioning a GP on examples, produces another
GP whose mean and covariance functions can be written in closed form.  Letting
$\vx_{t} = (x_1, \ldots, x_t)^\top$ be the instances received so far and
$\vy_{t} = (y_1, \ldots, y_t)^\top$ their ``scores'' $y_t = f(x_t)$
(possibly perturbed by Gaussian noise), the mean and covariance
functions conditioned on $(\vx_{t}, \vy_{t})$ are:
\begin{align}
    \mu_t(x)
        & \; = \vk_t(x)^\top \Gamma_t \vy_t \label{eq:mean}
    \\
    k_t(x,x')
        & \; = k(x,x') - \vk_t(x)^\top \Gamma_t \vk_t(x') + \rho^2 \label{eq:covariance}
\end{align}
Here we used $\vk_t(x) = (k(x_1, x), \ldots, k(x_t, x))^\top$,
$K_t = [k(x, x') \,:\, x, x' \in \vx_{t}]$,
$\Gamma_t = (K_t + \rho^2 I)^{-1}$,
and $\rho$ a smoothing parameter that models noise.
Given a GP with parameters $(\mu, k)$ and $x$, the value of $f(x)$ is
normally distributed with mean $\mu(x)$ and variance $k(x, x)$.  Hence,
the probability that $f(x)$ is non-negative is:
\[
    \textstyle
    \prob(f(x) \ge 0 \,|\, x) = \Phi\left( \frac{\mu(x)}{\sigma(x)} \right)
    \label{eq:probone}
\]
where $\Phi$ denotes the cdf of a standard normal distribution and
$\sigma(x) = \sqrt{k(x,x)}$.  This quantity is often used in classification tasks
to model the probability of the positive class, that is, $\prob(1 \,|\, x) = \prob(f(x) \ge 0 \,|\, x)$, see~\cite{kapoor2007active}.

\subsection{Incremental Multi-class GPs}

Incremental multi-class GPs (IMGPs) generalize Gaussian Processes to multi-class
classification~\cite{lutz2013want}.  An IMGP
can be viewed as a collection of GPs, one for each observed class $\ell \in \calY_t$, which share
the same precision matrix $\Gamma_t$ but have separate label vectors $\vy_{\ell,t}$.
The label vectors use a one-versus-all encoding:  an element of $\vy_{\ell,t}$ is $1$ if the
label of the corresponding example is $\ell$ and $0$ otherwise.  The
posterior mean function of the $\ell$-th GP is:
\begin{align}
    \mu_{\ell,t}(x)
        & \; = \vk_t(x)^\top \Gamma_t \vy_{\ell,t} \label{eq:meanmc}
\end{align}
Since the covariance function does not depend on the labels, it
remains the same as in Eq.~\ref{eq:covariance}.  The multi-class posterior
is obtained by combining the GP posteriors with a soft-max:
\[
    \textstyle
    \prob(\ell \,|\, x_t) = \frac{1}{Z} \exp \prob_\ell(1 \,|\, x_t), \; Z = \sum_{\ell'} \exp \prob_{\ell'}(1 \,|\, x_t)
    \label{eq:probl}
\]
Here $\prob_\ell(1 \,|\, x_t)$ is the posterior of the $\ell$-th GP
(Eq.~\eqref{eq:probone}) and $Z$ is a normalization factor.

IMGPs offer two major advantages.  First, in IMGPs the predictive variance is \emph{guaranteed} to increase with the distance from the training set, as illustrated by Figure~\ref{fig:example} (right).  This prevents IMGPs from being over-confident about classes and instances that differ significantly from its previous experience, a key feature when learning in the wild.
Another benefit is that IMGPs support incremental updates, i.e., in each iteration the updated precision matrix $\Gamma_{t+1}$ is computed from $\Gamma_t$ by exploiting the matrix-inversion lemma, without any matrix inversion~\cite{lutz2013want}.  This makes IMGPs scale much better than non-incremental learners and GPs;  see Section~\ref{sec:plusminus} for a discussion.

\begin{algorithm}[tb]
    \caption{Pseudo-code of \method. $\calY_0$ is provided as input.  All branches are stochastic, see the relevant equations.}
    \label{alg:isgp}
    \begin{algorithmic}[1]
        \For{$t = 1, 2, \ldots$}
            \State receive $x_t$ \label{eq:xt}
            \State $\predy_t \gets \argmax_{y \in \calY_{t-1}} \; \mu_{y}(x_t)$ \label{eq:pt} \Comment{Eq.~\eqref{eq:inference}}
            \If{uncertain about $\predy_t$} \label{eq:activebranch} \Comment{Eq.~\eqref{eq:activemath}}
                \State request label, receive $\noisyy_t$ \label{eq:activequery}
                \If{skeptical about $\noisyy_t$} \label{eq:skepticalbranch} \Comment{Eq.~\eqref{eq:skepticalmath}}
                    \State challenge user with $\predy_t$, receive $\consy_t$ \label{eq:skepticalquery}
                \Else
                    \State $\consy_t \gets \noisyy_t$
                \EndIf
                \State add $(x_t, \consy_t)$ to data set and update IMGP \label{eq:update}
                \State $\calY_t \gets \calY_{t-1} \cup \{\consy_t\}$
            \EndIf
        \EndFor
    \end{algorithmic}
\end{algorithm}

\subsection{\acronym}

We are now ready to present \method;  the pseudo-code is listed in Algorithm~\ref{alg:isgp}.
In each iteration $t$, the learner receives an instance $x_t$ and predicts the
most likely label (line~\ref{eq:pt}):
\begin{align}
    \textstyle \predy_t
        & \textstyle = \argmax_\ell \prob(\ell \,|\, x_t) = \argmax_\ell \frac{1}{Z} \exp \prob_\ell(1 \,|\, x_t)
        \nonumber
    \\
        & \textstyle = \argmax_\ell \Phi\left( \frac{\mu_{\ell,t}(x)}{\sigma_t(x)} \right) = \argmax_\ell \mu_{\ell,t}(x_t)
        \label{eq:inference}
\end{align}
where $\ell \in \calY_{t-1}$.  The last step holds because $\Phi$ is monotonically increasing and $\sigma_t(x)$ does not depend on $\ell$.

At this point, \method has to decide whether to request the label of $x_t$ (line~\ref{eq:activebranch}).  In line with approaches to selective sampling~\cite{cesa2006worst,beygelzimer2008importance}, \method prioritizes requesting the labels of uncertain instances, as these are more likely to impact the model.  This also limits the labeling cost as the model improves.  Intuitively, $x_t$ is uncertain if either $\mu_{\predy_t}(x_t)$ is small or $\sigma_t(x_t)$ is large;  in either case, Eq.~\eqref{eq:probone} ensures that $P_{\predy_t}(1 \,|\, x_t)$ is small.  Hence, \method queries the annotator with probability $\prob_{\predy_t}(0 \,|\, x_t)$.  This is achieved by sampling $a_t$ from a Bernoulli distribution with parameter $\alpha_t$, defined as:
\begin{align}
    \textstyle
    \alpha_t
        & = \prob_{\predy_t}(f(x_t) \le 0 \,|\, x_t) = 1 - \Phi\left( \mu_{\predy_t,t}(x_t) / \sigma_t(x_t) \right) \label{eq:activemath}
\end{align}
and querying the user if $a_t = 1$.  The choice is randomized so to prevent \method from trusting the model too much, which is problematic, especially during the first rounds of learning.  Randomization is a key ingredient in online learning and selective sampling, cf. \cite{cesa2006worst}.

If the check succeeds, \method has to decide whether to challenge the user's label (line~\ref{eq:skepticalbranch}).  If the user and the machine agree on the label\footnote{The original formulation of SKL tackles hierarchical multi-class classification, in which the user and the machine can agree on a parent of the prediction and the annotation.  For simplicity, we focus here on multi-class classification.  The pathological behavior of SKL that our method fixes affects the hierarchical setting, too.}, the probability of challenging the user should be small;  we set it to zero, for simplicity.  Otherwise, it should increase with $\prob_{\predy_t}(1 \,|\, x_t)$ and decrease with $\prob_{\noisyy_t}(1 \,|\, x_t)$.
Since these probabilities come from different GPs, a direct comparison is not straightforward.  In order to facilitate this, \method treats the GPs as if they were independent.  Under this modeling assumption, letting $f_\ell$ be a sample from the $\ell$-th GP, $\prob(f_{\predy_t}(x_t) \ge f_{\noisyy_t}(x_t))$ is a normal distribution with mean $\delta_t(x) = \mu_{\predy_t}(x) - \mu_{\noisyy_t}(x)$ and variance $\sigma_t(x)$.  \method determines whether to challenge the user by sampling from a Bernoulli with parameter $\gamma_t$:
\begin{align}
    \textstyle
    \gamma_t
        & = \prob(f_{\predy_t}(x_t) - f_{\noisyy_t}(x_t) \ge 0) = \Phi\left( \delta_t(x_t) / \sigma_t(x_t) \right) \label{eq:skepticalmath}
\end{align}
This is analogous to the case of active queries discussed above.  Despite relying on an (admittedly strong) modeling assumption, this strategy worked well in our experiments.

Once confronted by the learner, the user replies with a potentially cleaned label $\consy_t$.  As in the original formulation of SKL~\cite{zeni2019fixing}, this label is never contested by \method.  The reason is that in our target applications the user is collaborative and label noise is mostly due to temporary inattention.  Lastly, in line~\eqref{eq:update} the model is updated using the consensus example $(x_t, \consy_t)$ and the loop repeats.

\subsection{Advantages and Limitations}
\label{sec:plusminus}

\method improves on the original formulation of skeptical learning~\cite{zeni2019fixing} in several ways.  A major benefit is that IMGPs are never over-confident in regions far away from the training set.  This facilitates allocating the query budget and avoids pathological behaviors.  Our empirical analysis shows that the original formulation has no such guarantees.
\method is also simpler.  \method uses the IMGP itself to model the confidence in the annotator's label, whereas the original implementation relies on a separate model trained heuristically.  Also, learning is not heuristically split into stages and only two hyper-parameters are needed, namely $k$ and $\rho$.  Since hyper-parameters are hard to tune properly in interactive tasks, this is a substantial advantage.  The net effect is that \method performs better and more consistently.

A well-known weakness of GPs is their limited scalability, due to the need of storing all past examples and of performing a matrix inversion during model updates.  The latter is avoided here by using incremental updates, which reduce the per-iteration cost from $O(t^3)$ to $O(t^2)$.  This is enough for \method to run substantially faster than the original implementation of SKL and to handle weeks or months of interaction with no loss of reactivity, as shown by our real-world experiment.  Sparse GP techniques can speed up \method even further~\cite{quinonero2005unifying}.  Of course, GPs are not immediately applicable to lifelong tasks: these will require different (online) learning techniques.  However, lifelong classification is beyond the scope of this paper.

Another limitation of \method is that the active and skeptical checks (that is, Eqs.~\eqref{eq:activemath} and~\eqref{eq:skepticalmath}) rely on the GP of the predicted class only.  The active check can be easily adapted to use information from all classes known to the IMGP by replacing $\prob_\ell(1\,|\,x_t)$ with $\prob(\ell\,|\,x_t)$.  An adaptation of the skeptical check, however, is non-trivial and left to a future work.  In practice, this does not seem to be an issue, as \method works much better than the original implementation of SKL.

\section{Experiments}

We investigate the following research questions:
\begin{description}

    \item[\textbf{Q1}]  Does \method output better predictions than the original formulation of skeptical learning?

    \item[\textbf{Q2}]  Does \method correctly identify mislabeled examples?

    \item[\textbf{Q3}]  Does \method scale better than skeptical learning?

\end{description}
In order to address these questions, we implemented \method using Python 3 and compared it against three alternatives on a synthetic and a real-world data set.  The competitors are the original implementation of SKL~\cite{zeni2019fixing} based on random forests, denoted \rf, and two active IMGP baselines that never and always challenge the user, dubbed \gpnever and \gpalways, respectively.  The experiments were run on a computer with a 2.2 GHz processor and 16 GiB of memory.  The code and experimental setup can be downloaded from: \url{gitlab.com/abonte/incremental-skeptical-gp}.

\subsection{Synthetic Experiment}

\begin{figure*}[t]
    \centering
    \begin{tabular}{cccc}
    
        \multicolumn{2}{c}{
        \includegraphics[scale=.3]{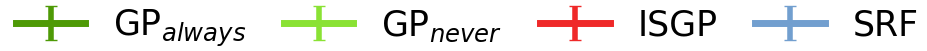}} &
        \multicolumn{2}{c}{
        \includegraphics[scale=.23]{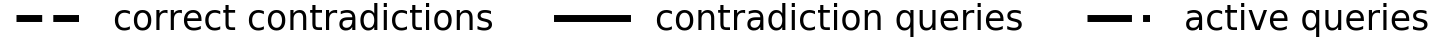}}
        \\
        
        \includegraphics[width=0.23\textwidth]{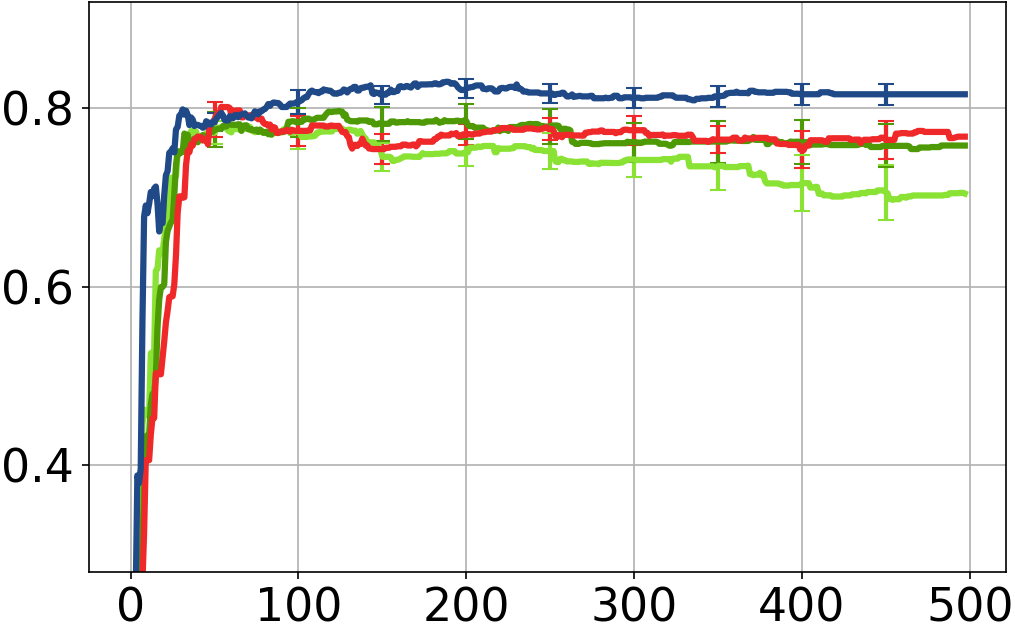} &
        \includegraphics[width=0.23\textwidth]{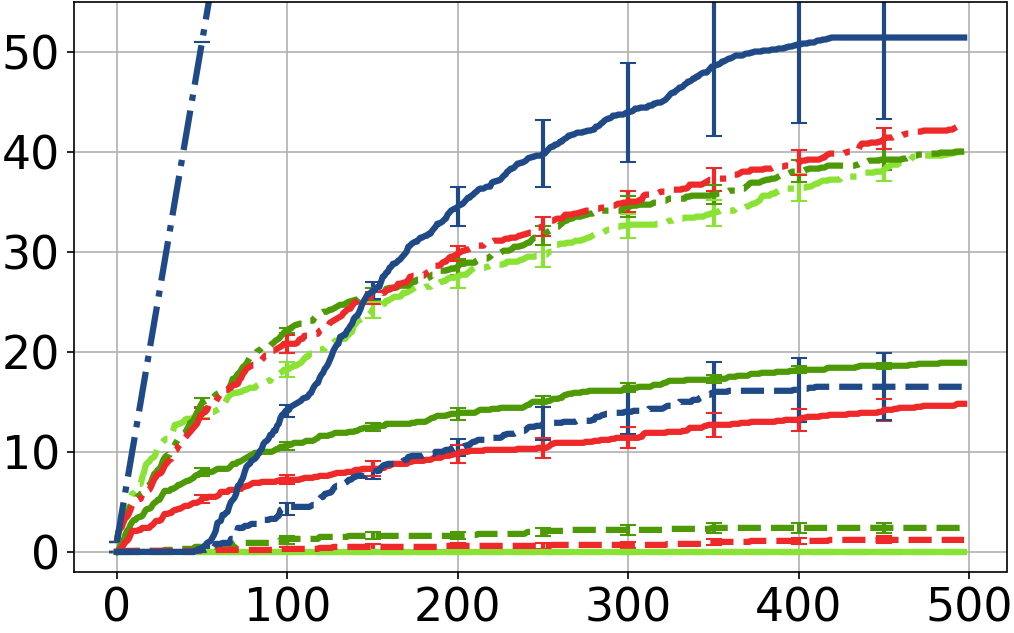} &
        \includegraphics[width=0.23\textwidth]{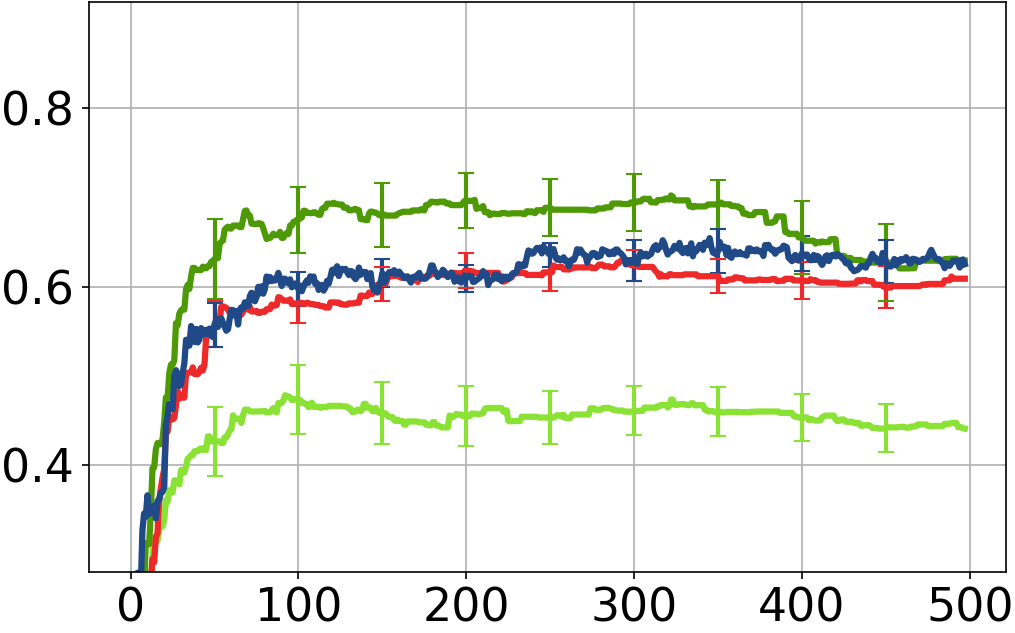} &
        \includegraphics[width=0.23\textwidth]{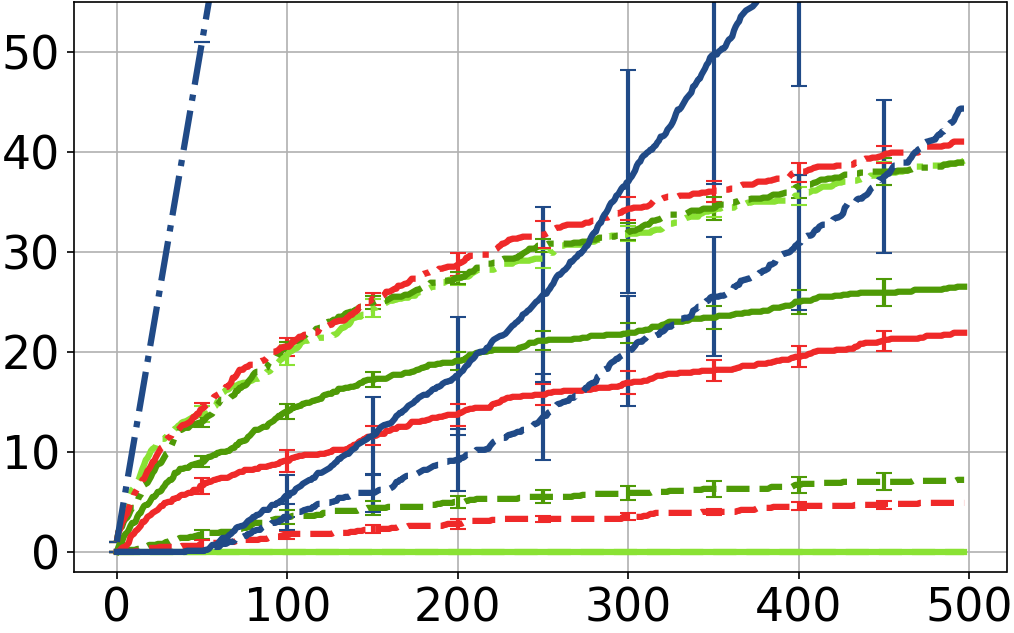} \\
        
        \includegraphics[width=0.23\textwidth]{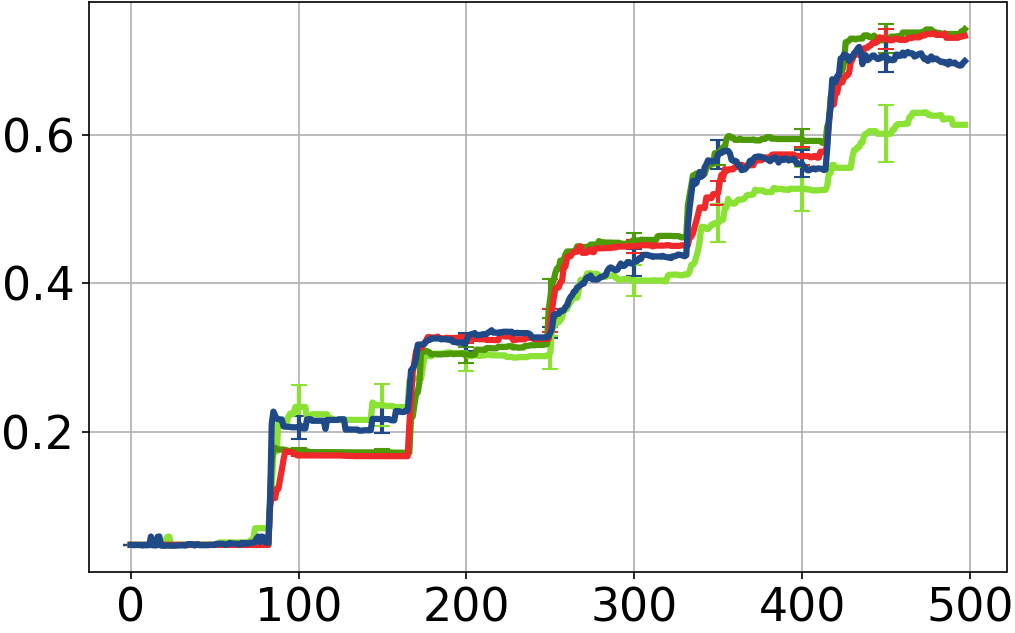} &
        \includegraphics[width=0.23\textwidth]{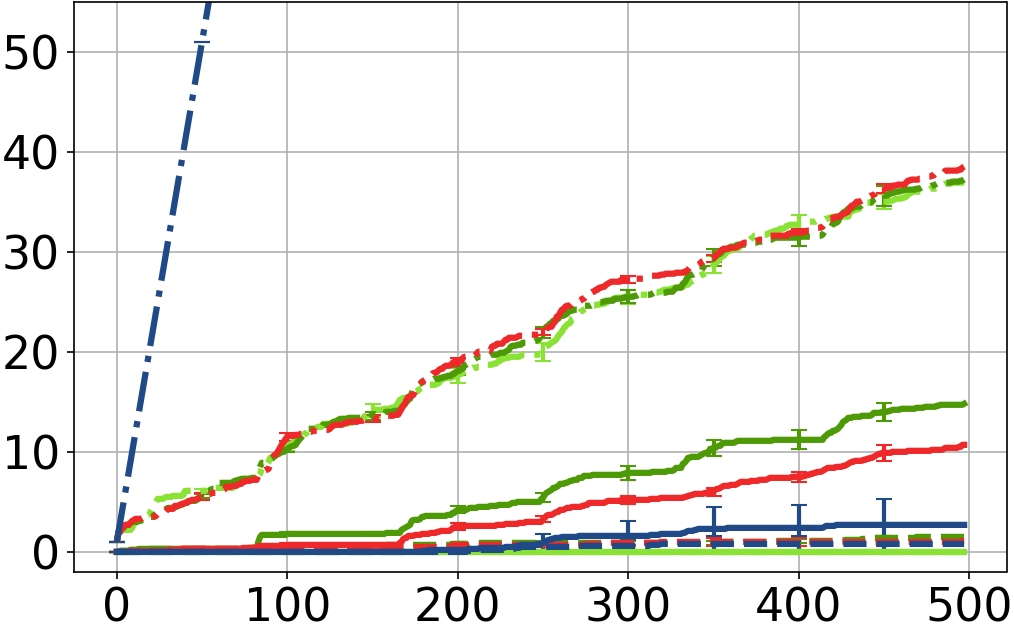} &
        \includegraphics[width=0.23\textwidth]{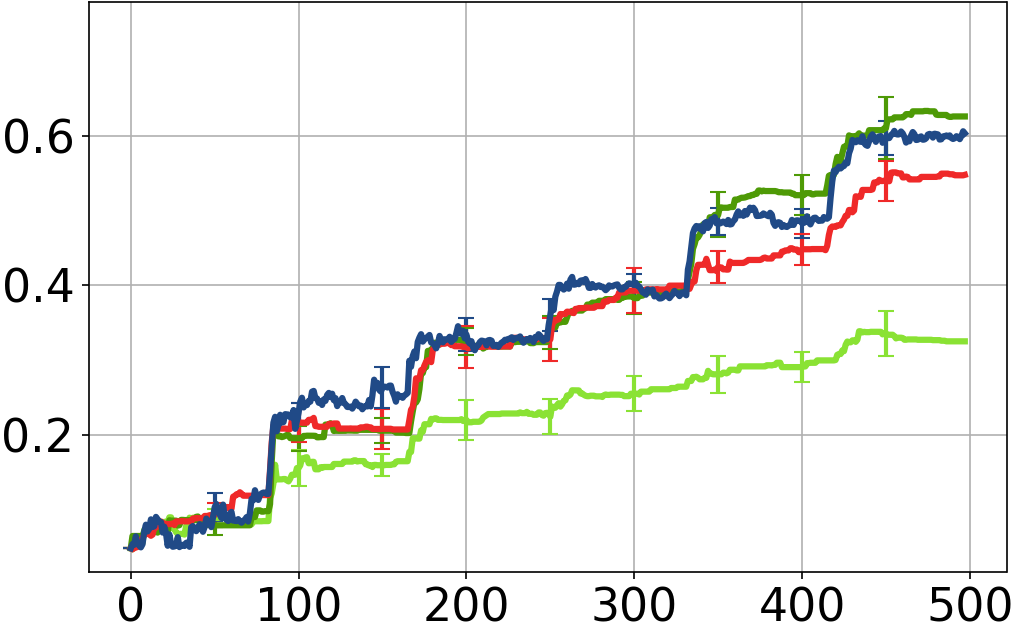} &
        \includegraphics[width=0.23\textwidth]{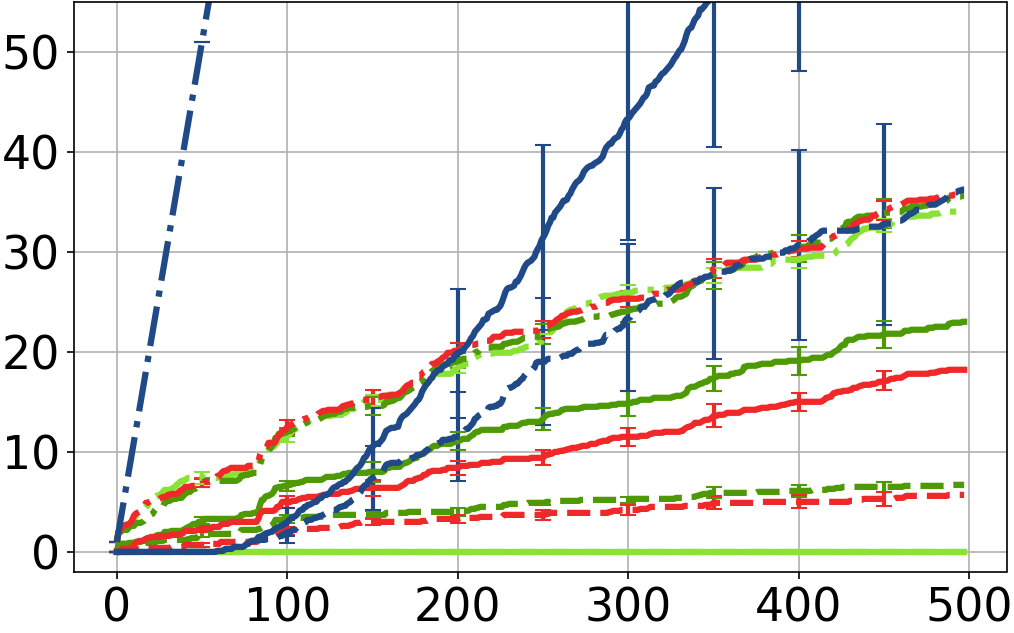} \\
        
        \includegraphics[width=0.23\textwidth]{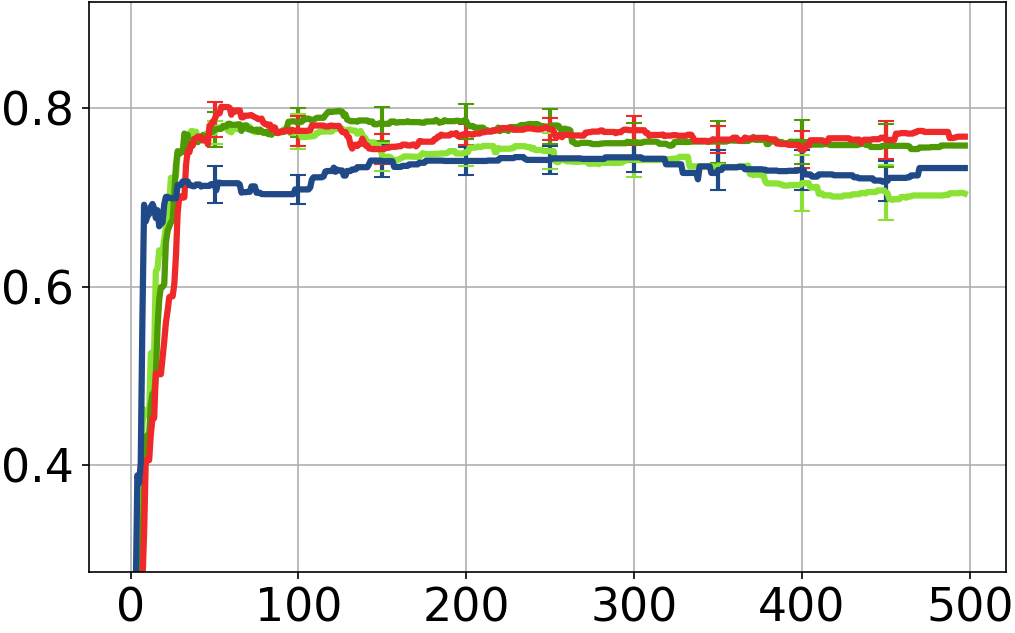} &
        \includegraphics[width=0.23\textwidth]{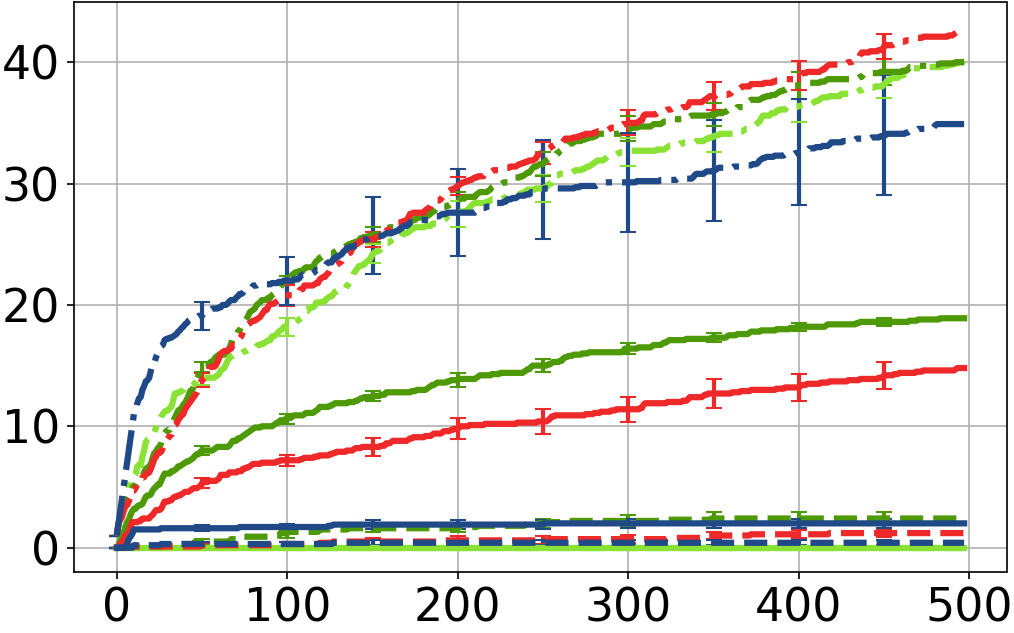} &
        \includegraphics[width=0.23\textwidth]{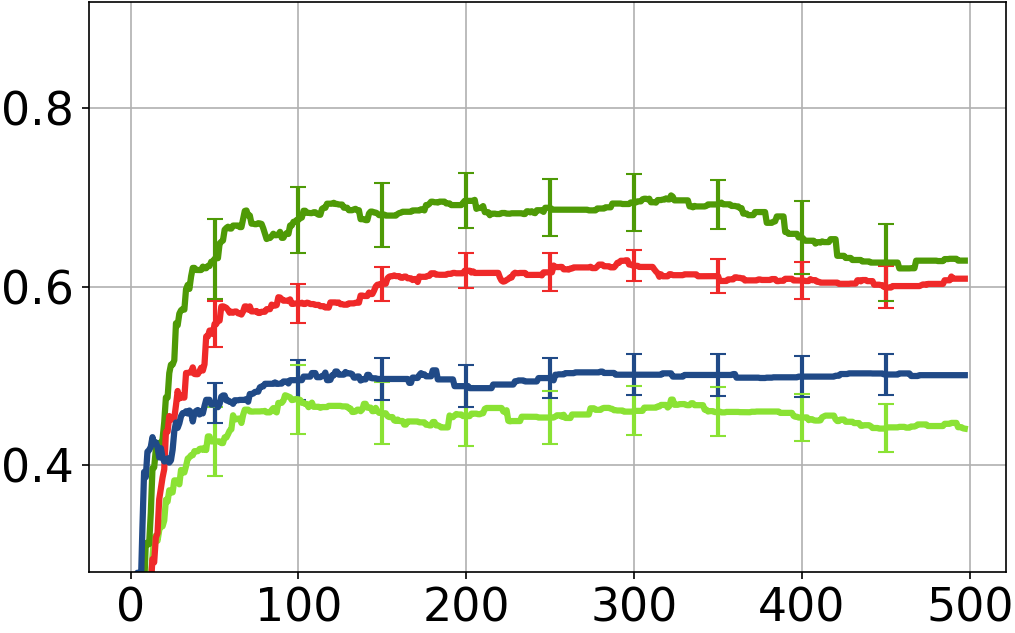} &
        \includegraphics[width=0.23\textwidth]{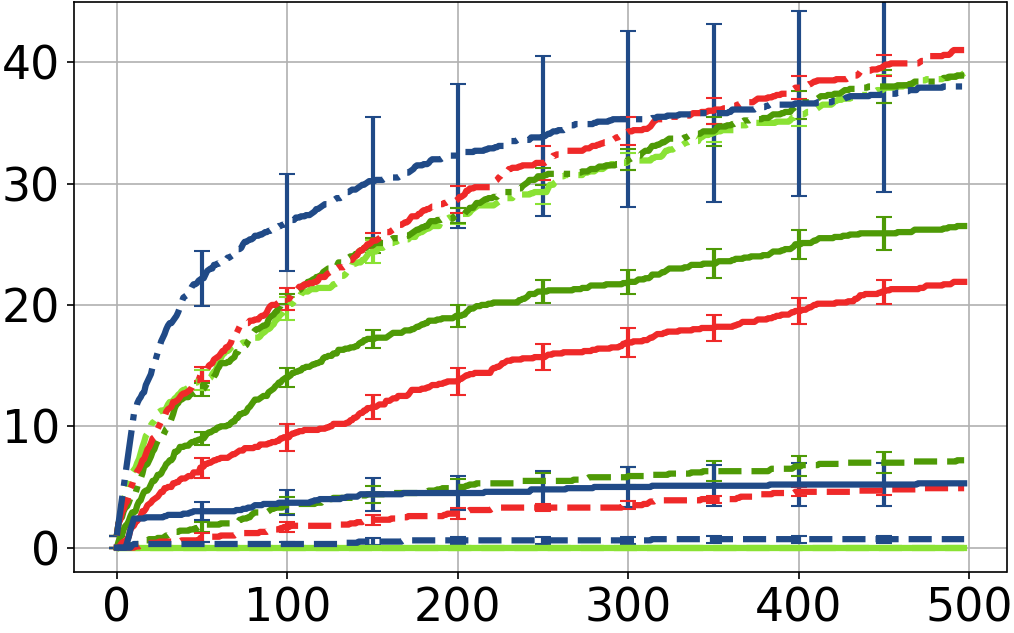} \\

        \includegraphics[width=0.23\textwidth]{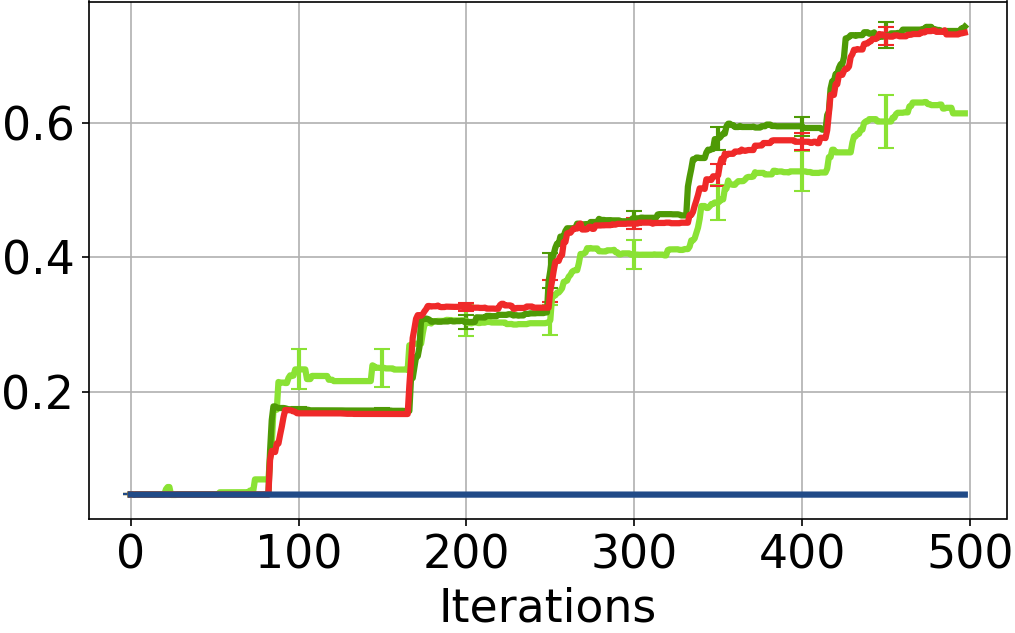} &
        \includegraphics[width=0.23\textwidth]{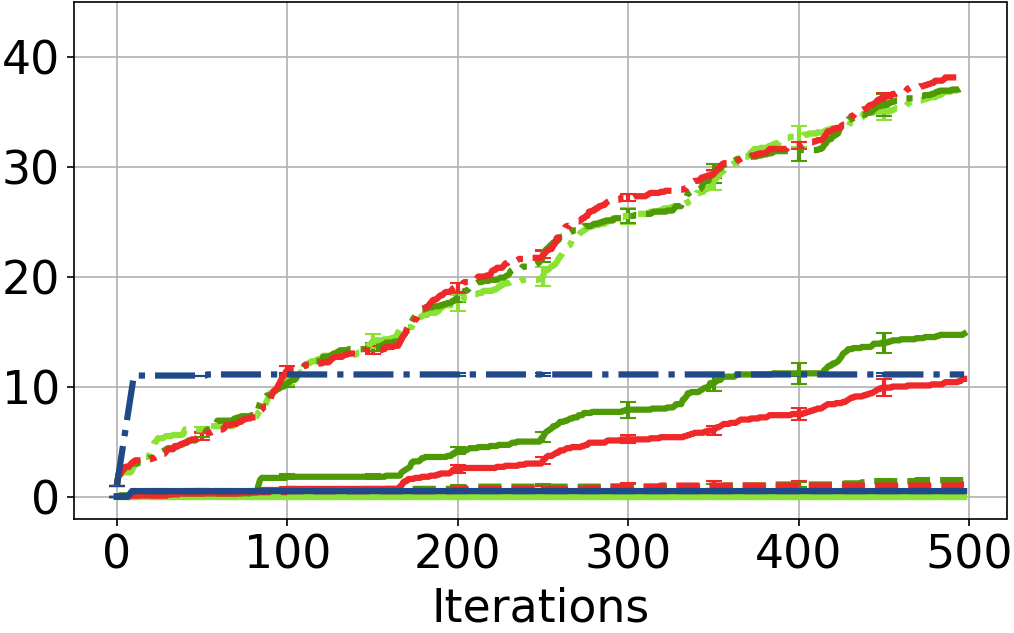} &
        \includegraphics[width=0.23\textwidth]{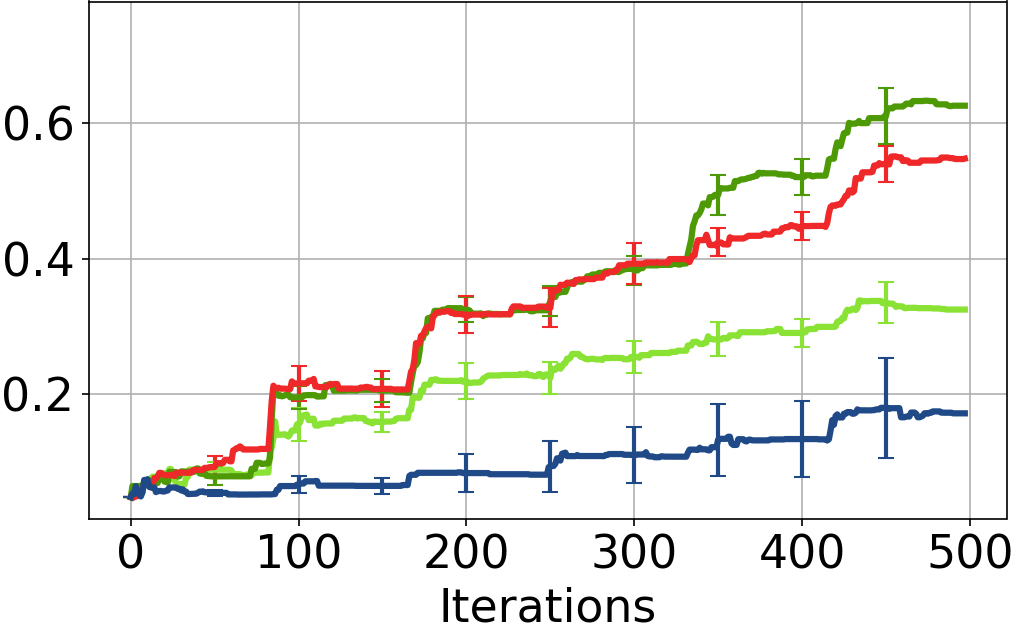} &
        \includegraphics[width=0.23\textwidth]{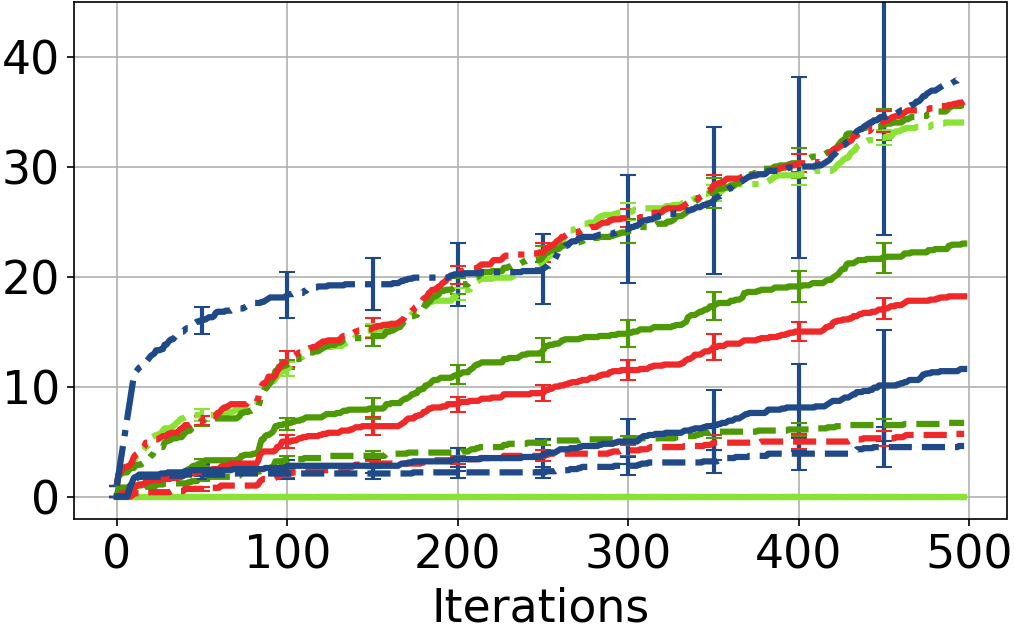}

    \end{tabular}
    \caption{\label{fig:synthetic}  Results on synthetic data.  The two leftmost columns report the $F_1$ and \# of labeling and contradiction queries (bars indicate the std. err.) for $10\%$ noise; rightmost columns do the same for $40\%$ noise.   Top two rows : \rf tuned to match $F_1$ of \method.  Bottom two rows:  \rf tuned to match \# of queries of \method (and forced to query at least $10$ labels).  Odd/even rows are random/sequential clusters, respectively.}
\end{figure*}

As a first experiment, we ran all methods on a synthetic data set with six classes, similar to Figure~\ref{fig:example}:  100 instances were sampled from six 2D normal distributions, one for each class, with different means and identical standard deviations (namely $1.5$).  As usual in active learning, the annotator's responses are simulated by an oracle.  Our oracle replies to labeling queries with a wrong label $\eta\%$ of the time.  We experimented with a low- ($\eta = 10$) and a high-noise regime ($\eta = 40$).  (Notice that $40\%$ noise rate is very high:  $50\%$ is the limit for learnability in binary classification~\cite{angluin1988learning}.)  While in~\cite{zeni2019fixing} the oracle always replies to contradiction queries with the correct label, our oracle answers with a wrong label $\eta\%$ of the time (unless the label being contested is correct, in which case no mistake is possible).  This is meant to better capture the behavior of human annotators, as the answer to contradiction queries can be incorrect.  Results obtained using the original oracle are not substantially different from the ones below.

All results are 10-fold cross validated.  For each fold, training examples are supplied in a fixed order to all methods.  The order has a noticeable impact on performance, so we studied two alternatives:  a)~instances chosen uniformly at random;  b)~instances chosen randomly from sequential clusters (red, then blue, \emph{etc.}).  This captures task shift, i.e., increasing number of classes.  $\calY_0$ matches the first example provided.  All GP learners used a squared exponential kernel with a length scale of $2$ and $\rho = 10^{-8}$, without any optimization.  The number of trees of \rf was set to\footnote{This matches the original paper.  With 100 trees, \rf is already more computationally expensive than \method, so we didn't increase it.} $100$.  The methods were evaluated based on their $F_1$ score and query budget usage.  For simplicity, the cost of skeptical queries was assumed to be similar to that of labeling queries.

\begin{figure*}[tb]
    \centering
    \begin{tabular}{cccc}
        \includegraphics[width=0.27\textwidth, height=0.15\textwidth]{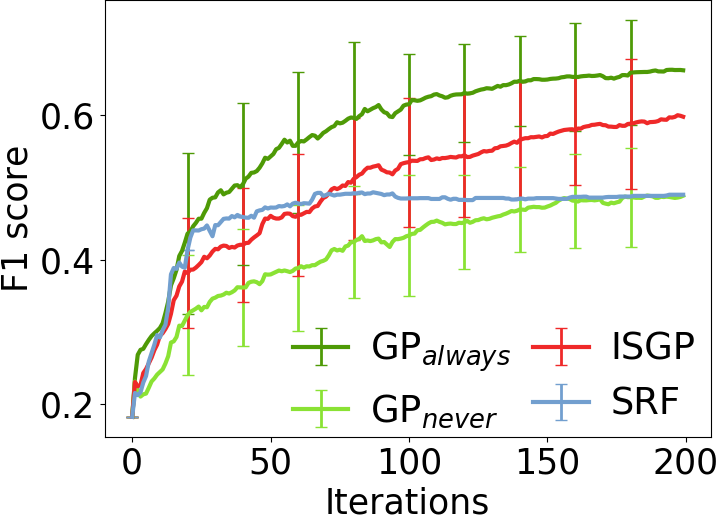} &
        \includegraphics[width=0.27\textwidth, height=0.15\textwidth]{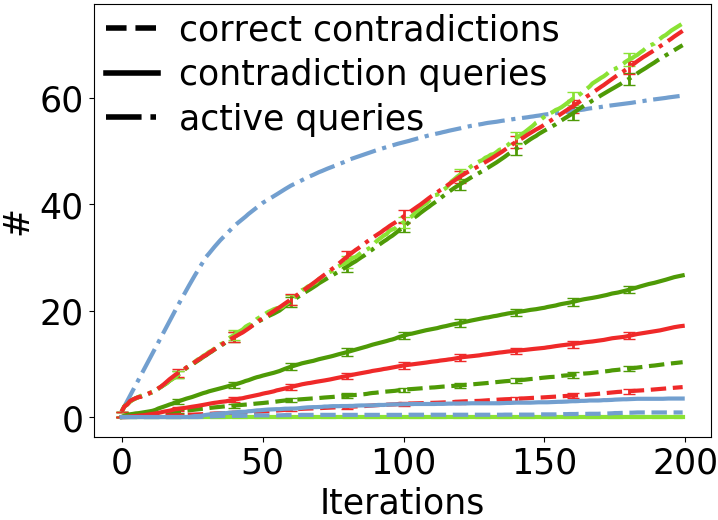} &
        \includegraphics[width=0.19\textwidth, height=0.15\textwidth]{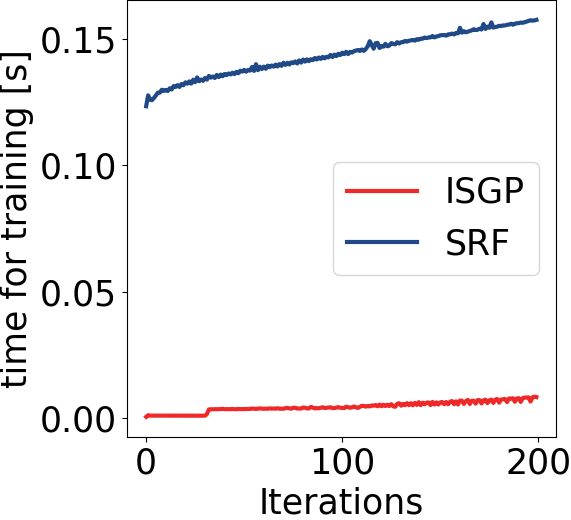} &
        \includegraphics[width=0.19\textwidth, height=0.15\textwidth]{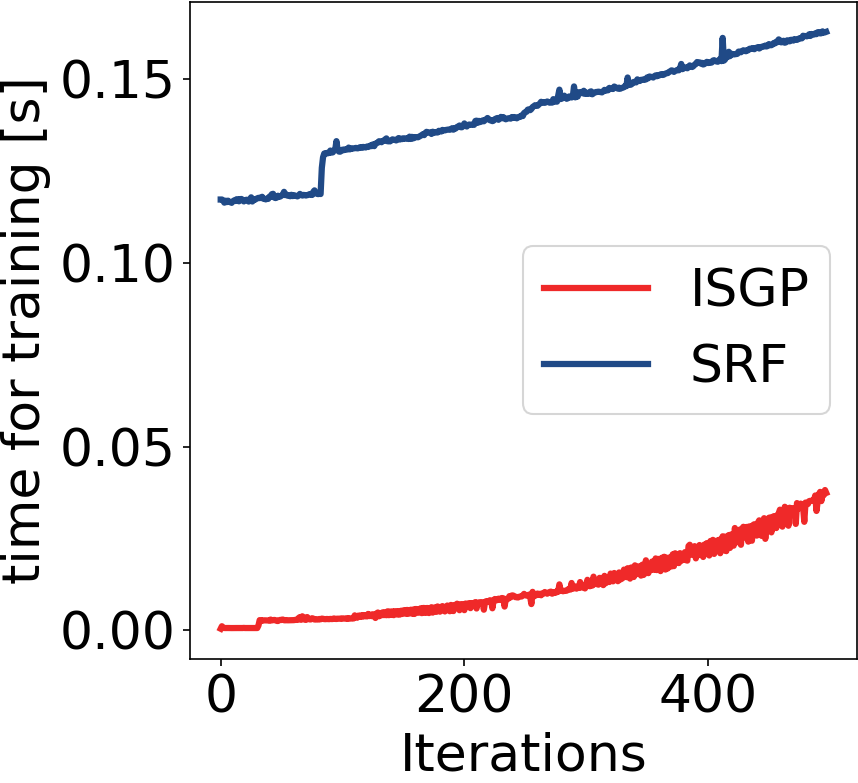}
    \end{tabular}
    \caption{\label{fig:exp_real_world} Results on location prediction.  Left to right: $F_1$ score, \# of queries (cumulative), and run-time (not cumulative) as learning proceeds on real-world and synthetic dataset (the training step is performed at each iteration).}
\end{figure*}

The cross-validated results can be viewed in Figure~\ref{fig:synthetic}.  The plots show the $F_1$ score on the left-out fold and the cumulative number of queries made (dash-dot line: active queries;  solid line: contradiction queries;  dashed line: contradiction queries that uncovered an actual mistake).  The two leftmost columns report the performance of all methods at $\eta = 10\%$ noise, the rightmost columns at $40\%$.  In order to enable a fair comparison, we tuned \rf to either match the $F_1$ score of \method or its query budget utilization.  This was achieved by tuning the hyper-parameter $\theta$ of \rf, which controls the length of the training and refinement stages of \rf, cf. Section~\ref{sec:skl}:  the longer the stages, the better the estimates acquired by the random forest but the worse the query usage.  These two settings are illustrated by the top two and bottom two rows of Figure~\ref{fig:synthetic}, respectively.  Finally, odd rows refer to random instance selection order, even rows to sequential order.

\rf worked well only in the low-noise, random order case (left columns, first row).  Here, it managed to outperform our method by about 5\%.  This setting, however, is not very representative of ICW, as the user is quite consistent and examples from all classes are quickly obtained.  In all other cases, \rf fails completely.  Two trends are clearly visible.  If tasked with reaching the $F_1$ score of \method, \rf tends to request the label of all new instances:  the blue curve increases linearly beyond the plot $y$ range.  This is because the value of $\theta$ needed to reach a high enough $F_1$ score also forces \rf to remain in refine and train stage for most iterations.  On the other hand, if the query budget is limited (bottom two rows), \rf quickly becomes over-confident and refuses to query the user.  This is especially troublesome with task shift (bottom row), as the random forest becomes confident after seeing examples from mostly one class, leading to abysmal performance.

Our method does considerably better.  Most importantly, \method does not suffer from pathological behavior and performs consistently across the board.  The $F_1$ score typically increases with the number of queries made, even in the high-noise scenarios, while querying is not too aggressive -- definitely not as aggressive as \rf.  The $F_1$ and query curves also show much lower variance compared to \rf in most cases, as shown by the narrower error bars.  It is easy to see that \method usually achieves $F_1$ score almost indistinguishable (in low-noise conditions, left two columns) or close (high-noise, right columns) to the $F_1$ of \gpalways with a comparable number of active queries and a smaller number of skeptical ones.  Moreover, \method always outperforms \gpnever in terms of $F_1$, as expected, while asking only $10$--$20$ extra queries.

\subsection{Location Prediction}

Next, we applied the methods to the location prediction task introduced in~\cite{zeni2019fixing},
which is reminiscent of our running example.  The data includes 20 billion readings from up to 30 sensors collected from the smartphones of 72 users monitored over a period of two weeks using a mobile app (I-Log~\cite{zeni2014multi}), for a total of 110 GiB.  The sensors are both hardware (i.g., gravity and temperature) and software (e.g., screen status, incoming calls).  The mobile app also asks every 30 minutes the user what he or she is doing, where and with whom.
We focus on location labels, for which an oracle exists capable of providing reliable ground truth annotations.  The task consists in predicting the location of the user as {\em Home}, {\em University} or {\em Others}.  The oracle identifies {\em Home} by clustering the locations labelled as home by the user via DBSCAN~\cite{ester1996density}, and choosing the cluster where she spends most of the time during the night.  {\em University} is identified using the maps of the University buildings, while all remaining locations are identified as {\em Others}.  Please see~\cite{zeni2019fixing} for the list of sensors and the pre-processing pipeline.
The GP-based methods use a combination of constant, rational quadratic, squared exponential and white noise kernels.  \rf uses $100$ decision trees, as in the synthetic experiments.

Figure~\ref{fig:exp_real_world} shows the result on the real-world dataset. The leftmost plot highlights the
$F_1$ scores and the subsequent one the number of queries. In this experiment, \rf is tuned to match the same number of queries of \method. The case where the two methods have a similar $F_1$ score is not reported since \rf shows the same behaviour as in the synthetic experiments (i.e., the number of active queries tends to increase rapidly).  As in the synthetic experiments, the predictive performance of \method lies between \gpalways and \gpnever, as expected.  The number of active queries is also in line with the baselines, while the number of skeptical queries is very limited, roughly $15$.  Notice that the $F_1$ of \rf plateaus at roughly 70 iterations, while the performance of \method keeps increasing up to iteration 200.  This trend is again explained by the fact that \rf becomes over-confident and requests the label of new instances very infrequently (second graph from left).   All in all, these results confirm the considerations made in the synthetic experiment on a more challenging real-world ICW task.
Finally, the rightmost graphs show the training times of \method and \rf respectively in the real-wold and the synthetic task. The advantage of the incremental updates is immediately apparent:  \rf is substantially more computationally expensive in both tasks, making it a poor candidate for ICW with thousands of data points.  Moreover, \method enjoys a reduction of about 70\% of the predicting time in the location prediction task (data not shown).

\section{Related Work}

Our work generalizes skeptical learning (SKL)~\cite{zeni2019fixing} to
incremental classification in the wild;  the relationship between
the two is analyzed in detail in Section~\ref{sec:plusminus}.

Two other related areas are open recognition and lifelong learning.
Open recognition (OR)~\cite{boult2019learning} refers to learning problems like face verification, in which not all classes are observed during training. The goal is to attain low risk also on the unknown classes~\cite{scheirer2013toward}.  To this end, the learner attempts to distinguish between instances that belong to known classes (for which a prediction can be made) and instances that do not.  This typically amounts to rejecting instances that lie away from the training set, thus bounding the chance of unjustified high-probability predictions~\cite{scheirer2014probability,rudd2017extreme,boult2019learning}. Generalizations prescribe to annotate the detected unknown-class instances and re-train the model accordingly~\cite{bendale2015towards} and to employ incremental learning~\cite{de2016online}, as we do.  ICW is not open in the above sense:  while the not all target classes are known, all incoming instances are \emph{labeled}.  What makes ICW hard is that the annotations are noisy, while OR is not concerned with shielding the model from noise.  An additional difference is that in ICW there is no distinction between training and testing stages, as prediction and learning are interleaved.  Moreover, skeptical learning requires and exploits interaction with human annotators, which is absent in OR.

In lifelong learning~\cite{thrun1996learning,baxter2000model} the learner witnesses a
sequence of different but correlated classification tasks and
the goal is to transfer knowledge from the previous tasks to the new ones. 
This is related to multi-task learning~\cite{skolidis2011bayesian,pillonetto2008bayesian}.
Surprisingly, most existing algorithms either assume a batch learning, although
some do support incremental or online learning;  cf. the discussion
in~\cite{denevi2018incremental}.  The main differences with ICW are that
lifelong learning is unconcerned with noise handling and it does not consider
interaction with human annotators.

\section{Conclusion}

We introduced interactive classification in the wild (ICW) and \method, a redesign of skeptical learning based on Gaussian Processes.  \method solves ICW while avoiding pathological scenarios in which the learner always or never queries the annotator.  Our empirical results showcase the benefits of our approach.

\section*{Acknowledgments}

The algorithm was substantially improved thanks to the input of the anonymous
referees.  The research of FG and AP has received funding from the European Union's Horizon 2020 FET Proactive project ``WeNet -- The Internet of us'', grant agreement No 823783.  The research of ST and AP has received funding from the ``DELPhi - DiscovEring Life Patterns'' project funded by the MIUR Progetti di Ricerca di Rilevante Interesse Nazionale (PRIN) 2017 -- DD n. 1062 del 31.05.2019.

\bibliographystyle{named}
\bibliography{paper}
\end{document}